\documentclass{article}

\usepackage[nonatbib,final]{neurips_2022}

\usepackage[utf8]{inputenc} 
\usepackage[T1]{fontenc}    
\usepackage{hyperref}       
\usepackage{url}            
\usepackage{booktabs}       
\usepackage{amsfonts}       
\usepackage{nicefrac}       
\usepackage{microtype}      
\usepackage{xcolor}         

\usepackage{graphicx}
\usepackage{longtable}

\title{Explicitly Increasing Input Information Density for Vision Transformers on Small Datasets}

\author{Xiangyu Chen\textsuperscript{\rm 1}, Ying Qin\textsuperscript{\rm 2,3}, Wenju Xu\textsuperscript{\rm 4}, Andrés M. Bur\textsuperscript{\rm 5}, Cuncong Zhong\textsuperscript{\rm 1}, Guanghui Wang\textsuperscript{\rm 6}$^*$\\
\textsuperscript{\rm 1}Department of EECS, University of Kansas, KS, USA. \\ 
\textsuperscript{\rm 2}Institute of Information Science, Beijing Jiaotong University, Beijing, China. \\
\textsuperscript{\rm 3}Beijing Key Laboratory of Advanced Information Science and Network Technology, Beijing, China. \\
\textsuperscript{\rm 4}OPPO US Research Center, InnoPeak Technology Inc, Palo Alto, CA, USA. \\
\textsuperscript{\rm 5}Department of Otolaryngology–Head and Neck Surgery, University of Kansas, Kansas City, KS, USA.\\
\textsuperscript{\rm 6}Department of CS, Toronto Metropolitan University, Toronto, ON, Canada. \\
\tt\small xychen@ku.edu, wangcs@ryerson.ca (* corresponding author)
}

\begin{document}

\maketitle

\begin{abstract}
Vision Transformers have attracted a lot of attention recently since the successful implementation of  Vision Transformer (ViT) on vision tasks. With vision Transformers, specifically the multi-head self-attention modules, networks can capture long-term dependencies inherently. However, these attention modules normally need to be trained on large datasets, and vision Transformers show inferior performance on small datasets when training from scratch compared with widely dominant backbones like ResNets. Note that the Transformer model was first proposed for natural language processing, which carries denser information than natural images. To boost the performance of vision Transformers on small datasets, this paper proposes to explicitly increase the input information density in the frequency domain. Specifically, we introduce selecting channels by calculating the channel-wise heatmaps in the frequency domain using Discrete Cosine Transform (DCT), reducing the size of input while keeping most information and hence increasing the information density. As a result, $25\%$ fewer channels are kept while better performance is achieved compared with previous work. Extensive experiments demonstrate the effectiveness of the proposed approach on five small-scale datasets, including CIFAR-10/100, SVHN, Flowers-102, and Tiny ImageNet. The accuracy has been boosted up to 17.05$\%$ with Swin and Focal Transformers. Codes are available at https://github.com/xiangyu8/DenseVT.
\end{abstract}
\section{Introduction}

{ResNet \cite{he2016deep} and its variants have been widely used in computer vision tasks in the recent decade due to its superior performance from introducing inductive bias \cite{cen2021deep,patel2022discriminative,yang2022unsupervised}. However, convolutional layers on the other hand limit its ability to capture global relations.} 
{Vision Transformer (ViT) \cite{dosovitskiy2020image}, as an emerging alternative backbone to ResNet in computer vision, expands the network  representation capacity by including global attention with multi-head self-attention modules, which shows better performance compared with ResNet in large-scale datasets like ImageNet \cite{deng2009imagenet} after pretraining on larger datasets JFT-300M \cite{sun2017revisiting}. However, the lack of locality in ViT results in worse generalization on small datasets compared with ResNet \cite{chenimproving,lee2021vision,gani2022train}. To release its dependency on more data and accelerate convergence, researchers try to add convolutional layers to transformers \cite{graham2021levit,guo2021cmt,wu2021cvt,xiao2021early}. {However, none of them solve this by increasing input density directly.}}

{As noticed in \cite{he2022masked}, transformers originate from Natural Language Processing and the main difference between languages and images is information density. This reminds us to increase the token information density of image input to close the gap and better generalize vision transformers on image tasks.}
{To achieve this, the principle of image compression can be utilized. Image compression is a widely explored task, aiming at transforming images from spatial domain to another domain to reduce storage and transmission costs, while keeping as much information as possible. JPEG is one of the popular Discrete Cosine Transform (DCT) based image compression algorithms, keeping about 95$\%$ information while high compressibility by reducing a large amount of less informative high-frequency details in its quantization process, increasing the information density in the frequency domain. Inspired by this, we propose to feed vision transformers with dense information in the frequency domain to approach the dense word embedding in NLP. Specifically, we perform block DCT on YCbCr images and then remove up to 174 non-functional frequency channels (192 in total) based on the channel-wise heatmaps. {This channel selection increases input density greatly by decreasing the size of input greatly while keeping most information, making it closer to word embedding in NLP and better generalizing vision Transformers on small datasets. The source codes will be available on the author's website.}}

The major contributions of this paper are as below:
\begin{itemize}
    \item We first propose to \textbf{explicitly} increase the input information density in vision Transformers to close the gap between language and image input.
    \item To form the dense input, we design a simple yet effective channel-wise heatmap-based strategy to select useful DCT frequency channels. Compared with the static shape-based channel selection strategy in previous work, our selection keeps fewer channels while showing better results for Swin Transformer.
    \item Experiments on Tiny ImageNet \cite{le2015tiny}, CIFAR-10/100 \cite{krizhevsky2009learning}, Flowers-102 \cite{Nilsback08} and SVHN \cite{netzer2011reading} with backbones Swin Transformer \cite{liu2021swin} and Focal Transformer \cite{yang2021focal} demonstrate the effectiveness of learning in the DCT frequency domain with vision Transformers.

\end{itemize}

\section{Related Work}
{This section includes some most related work regarding vision transformers and deep learning in non-RGB domain.

{\bf Vision Transformers.}
Vision Transformers have received a lot of attention and applications in recent years \cite{gajurel2021fine,ma2021miti,patel2022aggregating,sajid2021audio}. However, the original Vision Transformer, ViT, requires a large amount of data to learn local properties. 
To handle this, some researchers manage to integrate convolutional layers in different modules of vision Transformers, e.g. token or backbones, to capture both local and global information of the image and accelerate convergence considering that token embeddings can be mapped into fixed positions of the input images. For instance, \cite{xiao2021early} replaced the patch-based token preprocessing with several small-size convolutional layers directly and found a better, faster, and more robust convergence compared with the original ViT. In addition to this, LeViT \cite{graham2021levit}, CMT \cite{guo2021cmt} and CeiT \cite{yuan2021incorporating} also used a convolution-based tokenization on top of transformer blocks. Others add convolutional layers in the middle of Transformer backbones, e.g. CvT \cite{wu2021cvt} employed squeezed convolutional projection in replace of linear projection to obtain keys/queries/values for multi-head attention in transformer blocks, LocalViT \cite{li2021localvit} added depth-wise convolution layer to the feed-forward module in transformer blocks and CoaT \cite{xu2021co} implemented a depth-wise convolution-based position encoding in the multi-head attention layer.
Another folder to accelerate convergence is to design pyramid-like structure modelling CNN backbones, including PVT \cite{wang2021pyramid}, Swin Transformer \cite{liu2021swin}, MViT \cite{fan2021multiscale}, NesT \cite{zhang2022nested} and LiT \cite{pan2021less}. However, these works focus on extracting diverse features based on the low information density images themselves and put feature extraction pressure on the networks. Instead, this work explicitly increases the input information density at the beginning, undertakes some responsibility from the networks, and fills the gap between languages and images.}  

{\bf Learning in non-RGB domain. }
{Although most computer vision algorithms are RGB image-based, there are some efforts to explore learning in other domains. One popular domain is from one stage in the JPEG encoding and decoding process. JPEG format is widely employed to compress images to save the cost of transmitting and storing images. In the encoding stage, it consists of converting RGB images to YCbCr, quantifying block DCT (where information loss happens), and encoding with Huffman. The decoding process is just the inverse. And normally, the widely used RGB image input is the output after decoding JPEG images. \cite{gueguen2018faster} proposes to learn from the quantified DCT domain in the decoding stage, i.e. feed the quantified DCT coefficients to the network directly before they are converted back to the RGB domain, accelerating convergence by removing the decompressing process to transform DCT representation back to image pixel and learn similar feature with its first layers. \cite{ehrlich2019deep} formulates deep learning in JPEG transform domain, representation after quantifying the DCT coefficient in the encoding stage, and shows the equivalence after conversing models learning on the RGB domain to this intermediate JPEG domain. \cite{ghosh2016deep} incorporates DCT transformation on the feature generated by the first convolutional layer, speeding up convergence a lot compared with using standard CNN. The most relevant work is \cite{xu2020learning}, which proposes to learn from an earlier stage, block DCT domain before quantization, and then disregard learnable useless frequency channels to save communication costs between CPU and GPU. However, \cite{xu2020learning} designs some learnable parameters and Gumble sampling to select channels while it turns out to be a square shape-based selection strategy works a little bit better. Differently, this work employs GradCAM heatmap to check functional channels directly, which is simpler, more intuitive and efficient.}

\section{Method}
\begin{figure*}[t]
\centering
\includegraphics[width=0.9\linewidth]{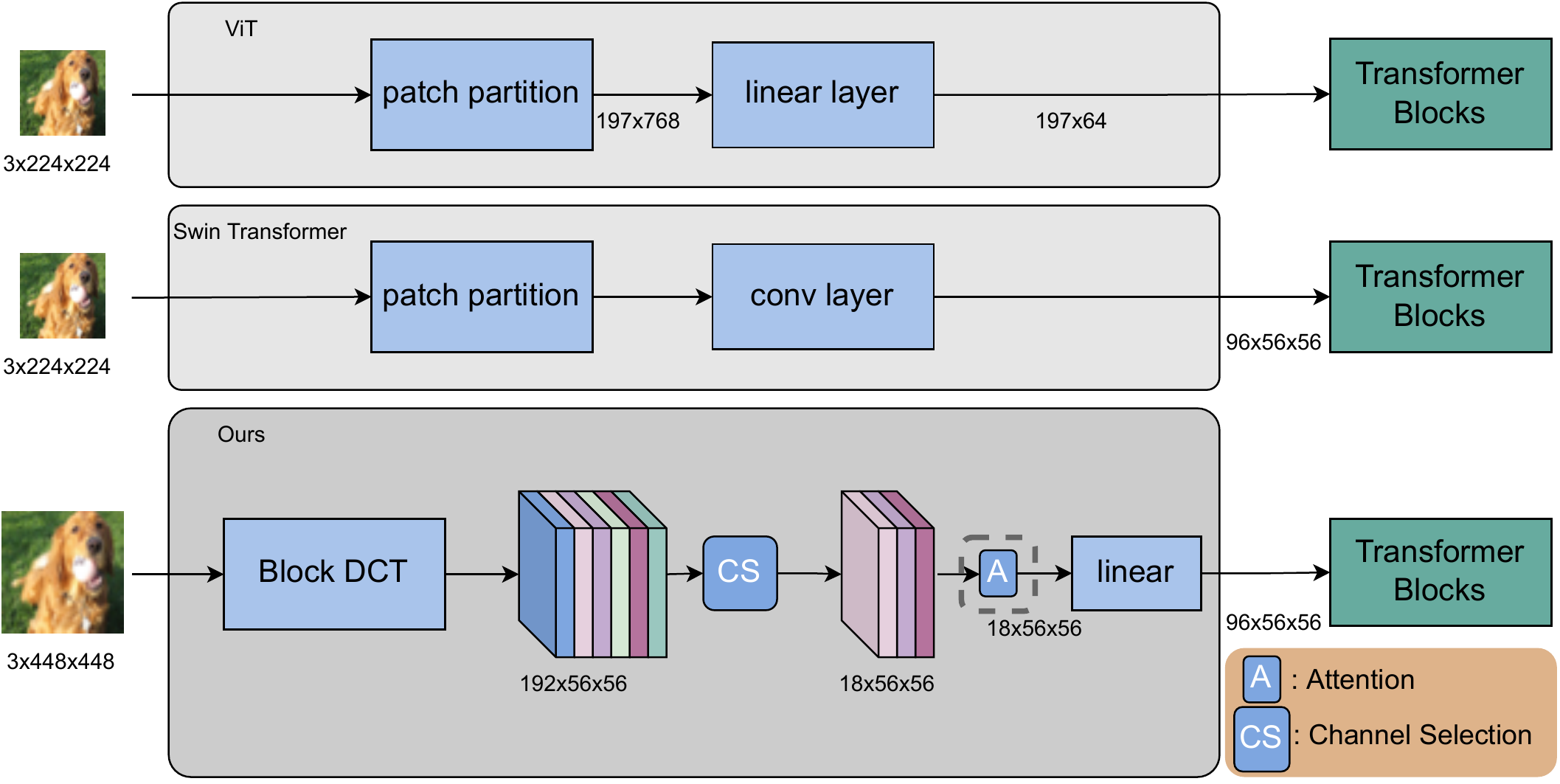}
\caption{Vision Transformer (ViT), Swin Transformer, and our Dense DCT-based vision Transformer. In our dense DCT pre-processing, RGB images are resized to 448$\times$448 so that the output size of DCT remains the same, i.e., 56$\times$56. Then, the frequency representation with all 192 frequency channels goes through their channel selection based on the heatmap results, and a dense frequency representation with only 18 channels is obtained. The attention module is optional for the dense frequency representation to pass through directly before the Transformer block. Finally, to keep the input channel and following Transformer blocks the same, we pass the 18-channel input to a linear layer and get 96 channels as in Swin Transformer.}
\label{fig:comparison}
\end{figure*}

{The proposed framework is shown in figure \ref{fig:comparison}. In this section, we will introduce the details of each preprocessing module, including the block DCT, channel-wise heatmap-based channel selection strategy, and the relationship between frequency channel selection and information density.}

\subsection{Block DCT}
\begin{figure*}
\centering
\includegraphics[width=1.0\linewidth]{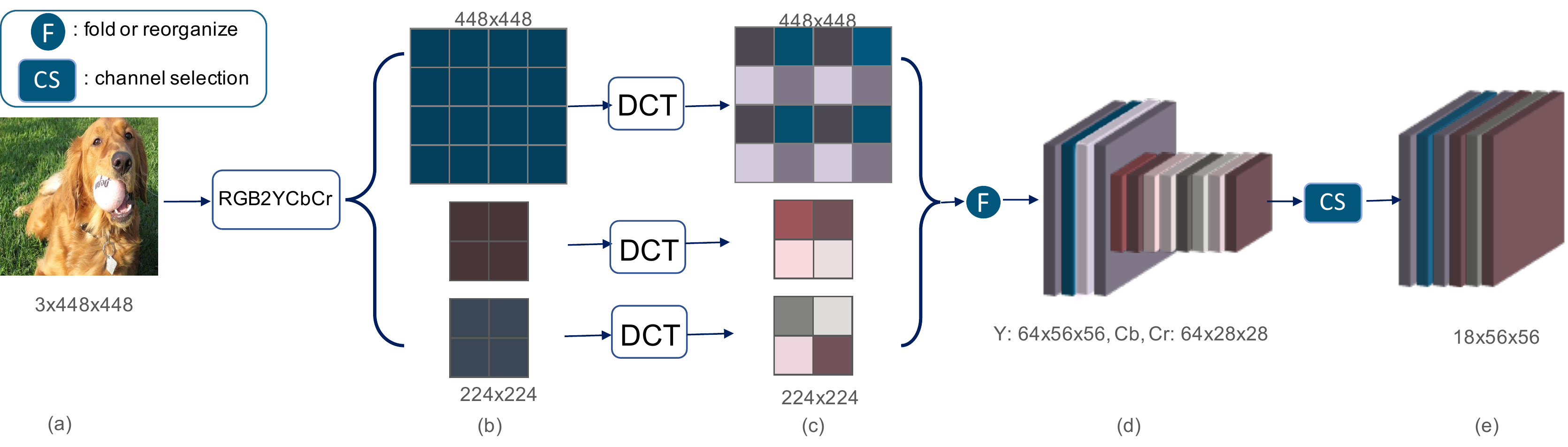}
\caption{Block DCT illustrated by 8$\times$8 DCT transformation. (a) The resized RGB images. (b) Convert RGB images to YCbCr with Cb and Cr channels downsampled. (c) Perform patch-based non-overlapping DCT for all 3 channels. (d) Rearrange the same frequency from all positions to the same channel. (e) Upsample Cr and Cb channels and then select frequency channels.}
\label{fig:dct}
\end{figure*}

{Block DCT is applied as in \cite{xu2020learning} and its illustration can be found in Figure \ref{fig:dct}. First, we convert the RGB images into YCbCr representation. Then, apply DCT transformation on each small block (e.g. $8\times8$) of the YCbCr component. After that, we reorganize DCT coefficients from the same frequency to one channel keeping block positions. Finally, we remove useless frequency channels after applying the channel selection strategy. As the human visual system is more sensitive to luminance component Y than the color components Cb and Cr, we downsample the Cb and Cr components as in JPEG compression in step (b) and then upsample them in step (d).}

\subsection{Channel-Wise Heatmap}
To learn better from the DCT frequency domain, \cite{xu2020learning} utilized a Gumbel softmax-based feature selection algorithm to reduce input channels while they found their shape-based (i.e., top left square) fine selection strategy achieves competing performance compared with the keeping learned frequency channels from feature selection algorithm. However, shape-based selection in the frequency domain lacks interpretability. Instead, we propose a channel-wise heatmap-based channel selection strategy as shown in \ref{fig:GradCAM}.

GradCAM \cite{selvaraju2017grad} is a method used to find the functional region of an image input in CNN by relating the output high-level features with their input positions. However, for input with multiple meaningful channels, different input channels might function differently as in \cite{xu2020learning}. To evaluate how each input contributes to the final high-level features, we mask out all channels except the candidate channel and then calculate the GradCAM heatmap, i.e., channel-wise heatmap. 

\begin{figure*}[t]
\centering
\includegraphics[width=0.7\linewidth]{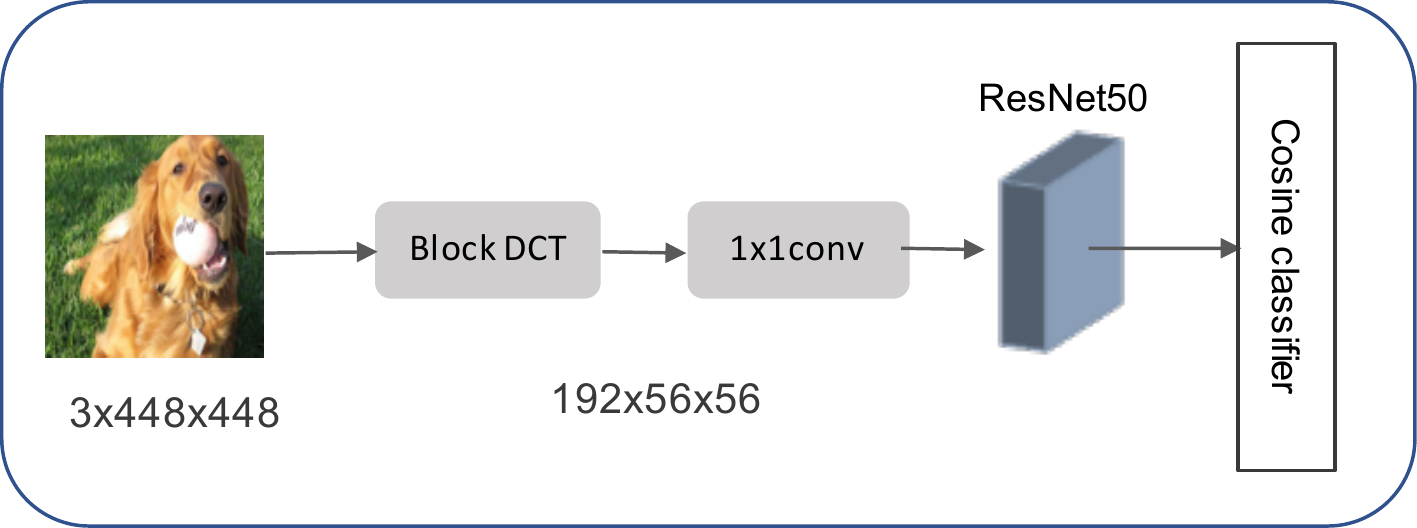}
\caption{Training pipeline to calculate the channel-wise heatmap.}
\label{fig:train}
\end{figure*}

\begin{figure*}[t]
\centering
\includegraphics[width=0.8\linewidth]{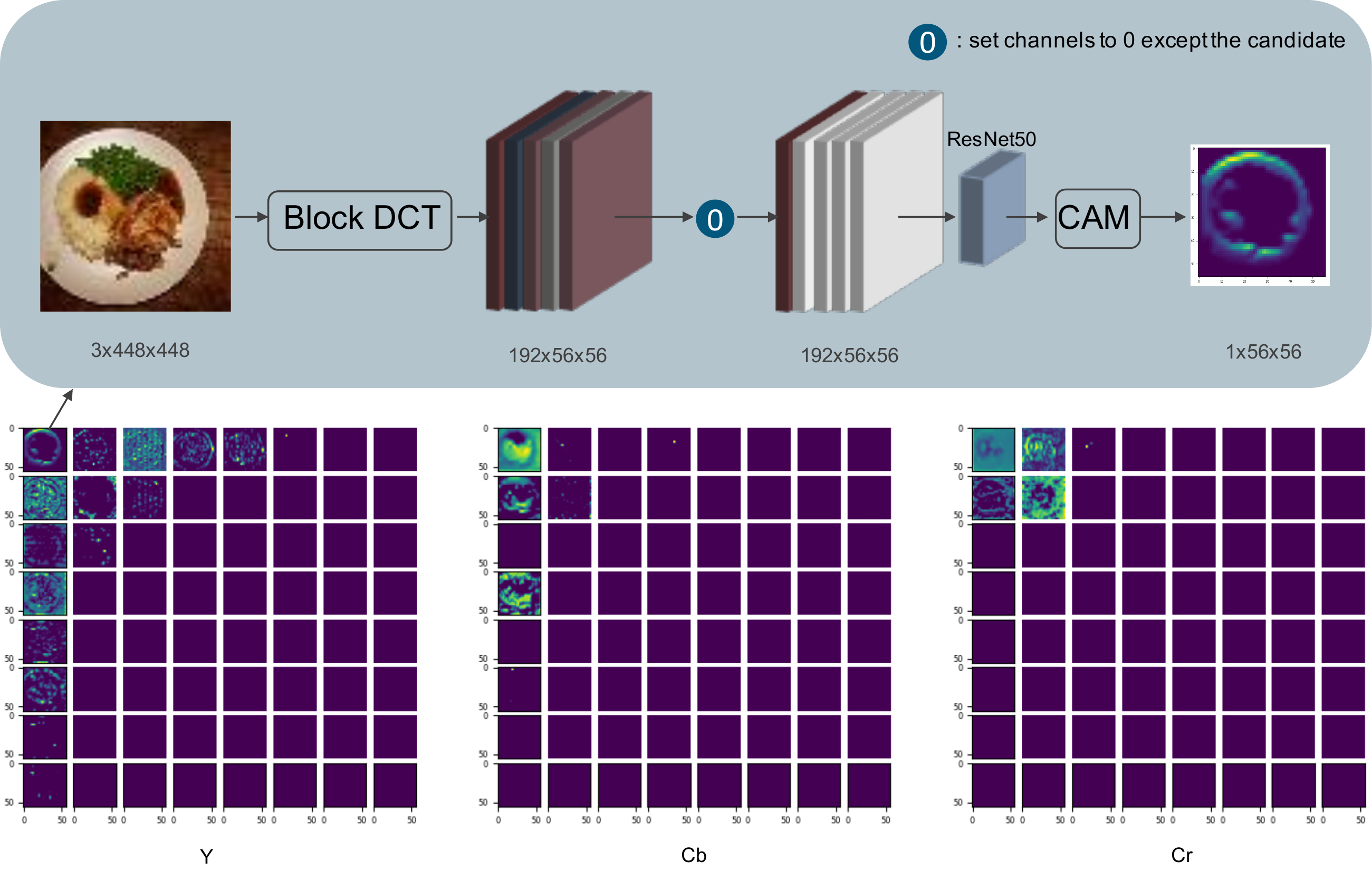}
\caption{Calculate the input channel-wise heatmap for one image based on pretrained model in \ref{fig:train}. 8$\times$8 DCT preprocessing on the RGB image results in a tensor with full 192 frequency channels. Then all channels except the candidate are set to zeros. The GradCAM heatmap is calculated based on the masked tensor and the model trained with full frequency channels and yields the input channel-wise heatmap for the candidate channel, e.g., channel 0 in this figure.}
\label{fig:GradCAM}
\end{figure*}

{Specifically, frequency representations with full frequency channels, i.e. 192 for $8\times8$, are fed into a ResNet50 backbone with cosine classifier as shown in Figure \ref{fig:train}. After that, to obtain the channel-wise heatmap, keep only one candidate frequency channel of the frequency representation and set other channels to 0 as shown in Figure \ref{fig:GradCAM}. Then, we pass the revised frequency representation into the pretrained model and calculate the GradCAM heatmap. Following the same process in Figure \ref{fig:GradCAM}, each image has 192 heatmaps (for $8\times8$ block DCT), and each heatmap is $56\times56$ following block position layout.}

\begin{figure*}[t]
\centering
\includegraphics[width=0.8\linewidth]{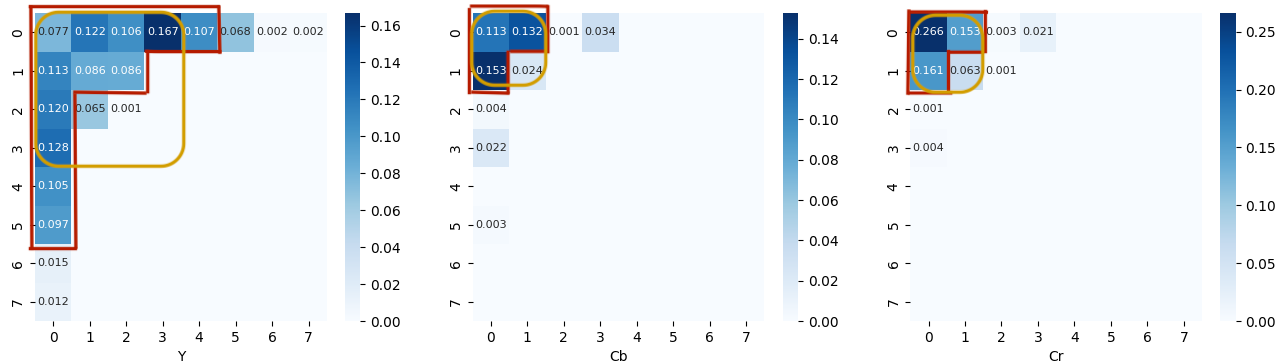}
\caption{Averaged channel-wise GradCAM heatmaps for $8\times 8$ by averaging all $56\times56$ for $8 \times 8$ positions in all training images from the same frequency channel. Channels in red are 18 selected channels using our method and those in yellow are 24 channels based on the method in \cite{xu2020learning}.}
\label{fig:sns_ResNet50}
\end{figure*}

{In this way, we calculate the average channel-wise heatmap of images from the training set and then average across all $56\times56$ positions for each channel heatmap. The results for $8\times8$ block  DCT can be found in Figure \ref{fig:sns_ResNet50}, where each frequency channel has only one value. This result is surprisingly consistent with the one from the channel selection network proposed in \cite{xu2020learning}, while our method requires no extra parameters. This indicates that the channel selection network in \cite{xu2020learning} is indeed learning functional channels as supposed. Finally, to select kept frequency channels and get dense DCT input, we simply set thresholds based on values in Figure \ref{fig:sns_ResNet50} and remove the channels that have no contribution to the performance. Frequency channels in red blocks are our 18 kept channels after comparing with the accuracy from different thresholds,  and those 24 channels within yellow squares are based on the square channel selection strategy concluded in \cite{xu2020learning}}.

\subsection{Frequency Channel Selection and Information Density}
\begin{figure}[h]
\centering
\includegraphics[width=0.4\linewidth]{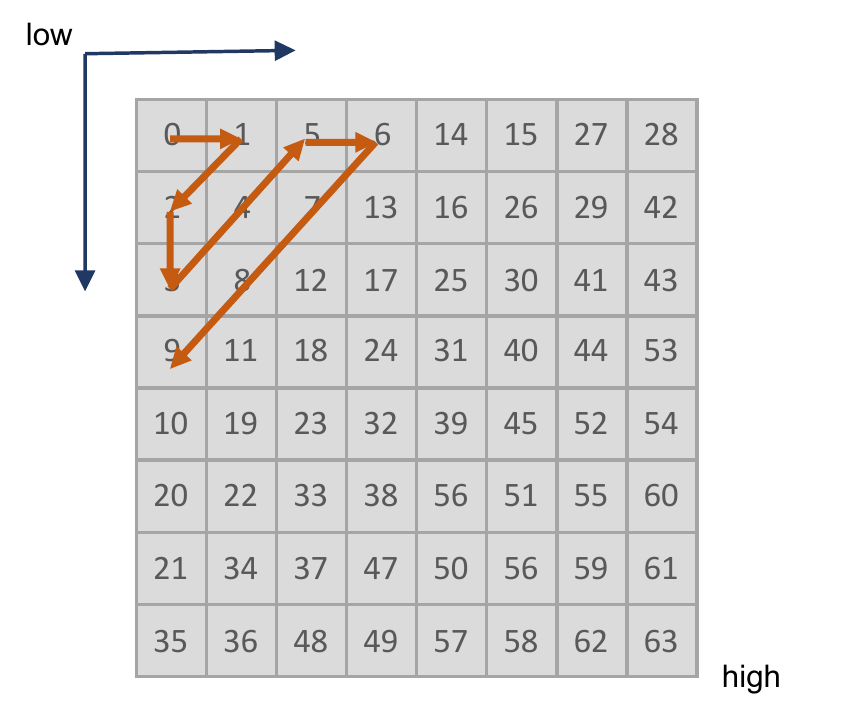}
\caption{Zig-zag route for $8\times 8$ DCT transformation.}
\label{fig:zigzag}
\end{figure}

\begin{figure}[h]
\includegraphics[width=\linewidth]{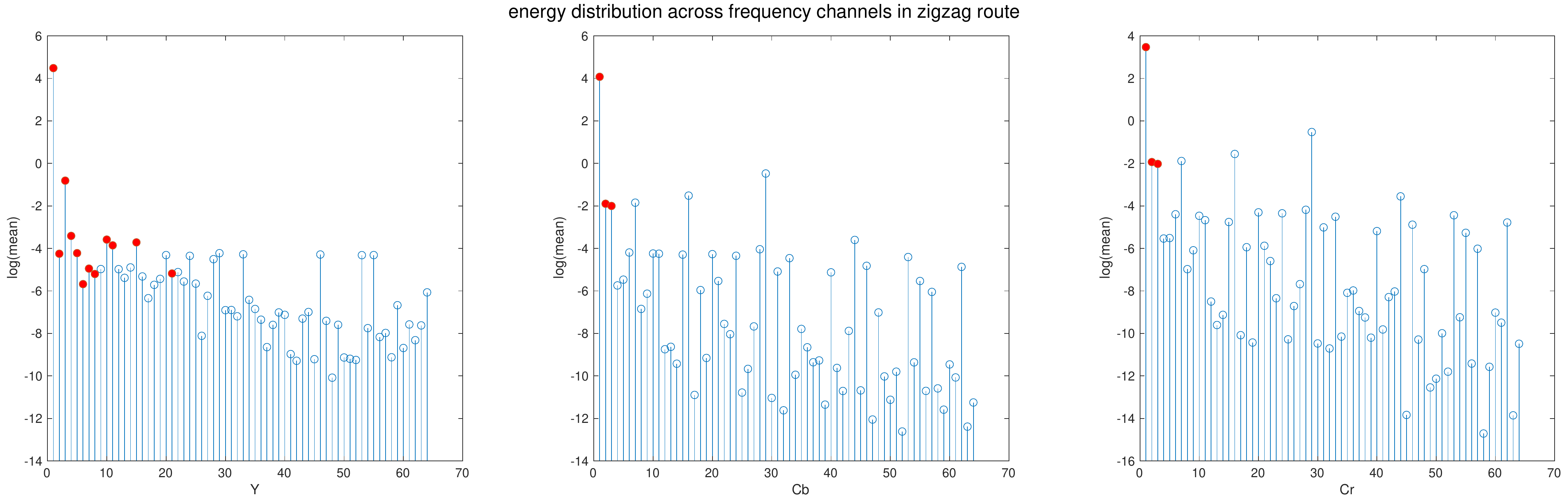}
\caption{Energy distribution across all 192 frequency channels from Y, Cb and Cr components.}
\label{fig:energy}
\end{figure}

{In DCT transformation, top left corner represents lower frequencies and bottom right are the higher ones. To visualize the energy distribution, we go through all frequency channels following the zig-zag route as shown in Figure \ref{fig:zigzag} and get Figure \ref{fig:energy}. Figure \ref{fig:energy} we know, most energies are pushed to sparse and low-frequency channels (top left on Figure \ref{fig:zigzag}) and most high-frequency channels have only few energy (which is also the principle for JPEG compression to disregard those zero channels after quantization). According to Figure \ref{fig:energy}, most selected channels (denoted by red dots) also lie on the low-frequency part. This indicates that, after channel selection, the overall information entropy is slightly reduced, while the size of the frequency representation is significantly decreased, i.e. the average information entropy or the information density is increased greatly.} 

\section{Experiments}
\subsection{Setup}
\paragraph{Dataset Information} 5 small datasets are employed to test the image classification performance of our framework, Tiny ImageNet \cite{le2015tiny}, CIFAR-10/100 \cite{krizhevsky2009learning}, Flowers-102 \cite{Nilsback08}, and SVHN \cite{netzer2011reading}. More details of these datasets are shown in Table \ref{table:dataset}. 
\begin{table}[t]
\centering
\caption{Dataset information for classification on small datasets}
\begin{tabular}{l|r|r|c|c}
\toprule\noalign{\smallskip}
 \bf{Dataset} & \bf{Train split} & \bf{Val split} & \bf{Size} & \bf{ Class \#} \\
\noalign{\smallskip}
\hline
  Tiny ImageNet& 100,000    & 10,000 & 64  & 200   \\
  CIFAR-10      & 50,000     & 10,000 & 32  & 10    \\
  CIFAR-100     & 50,000     & 10,000   & 32  & 100   \\ 
  SVHN        & 73,257     & 26,032   & 32  & 200   \\ 
  Flowers-102    & 2,040     & 6,149 & $>224$  & 102    \\
  \bottomrule
\end{tabular}
\label{table:dataset}
\end{table}
\setlength{\tabcolsep}{1.4pt}

\paragraph{Training Setup} Data augmentation methods we used include CutMix \cite{yun2019cutmix} and Mixup \cite{zhang2017mixup} as in \cite{liu2021efficient}. The initial learning rate is set to $5e^{-4}$ and 20 epochs warmup. We report the best top-1 accuracy after 100 epochs. The batch size is 128 per GPU and four A100 GPUs are used in total. Following \cite{liu2021efficient}, we resize all images to 3$\times$224$\times$224 on all datasets to get our baselines and resize all images to 3$\times$448$\times$448 unless otherwise specified. For a fair comparison, all vision Transformer backbones we selected have a comparable amount of parameters with ResNet50.

\subsection{Results}

\begin{table}
\centering
\caption{Top 1 classification accuracy on CIFAR-10/100, SVHN, Tiny ImageNet and Flowers-102 dataset}
\begin{tabular}{r|c|ccccc}
\toprule\noalign{\smallskip}
Backbone & Param.\# &CIFAR-10 & CIFAR-100 &SVHN &Tiny IMG & Flowers\\
\noalign{\smallskip}
\hline
 ResNet50\cite{he2016deep} & 25M & 92.77& 73.24 &96.78 &63.43& 41.34\\   
 Deit-S\cite{touvron2021training} & 22M & 81.79 & 59.11&90.64 &50.09&37.71\\ 
 T2T-ViT-14\cite{yuan2021tokens} &22M & 87.58 & 67.22&96.53&55.97&23.81 \\ \hline
 Swin-Tiny \cite{liu2021swin}   & 28M  & 81.91 &62.30&91.29&56.28&32.04\\
DenseDCT\_Swin-T& 28M   &   84.20 & 67.10 &  95.65& 57.36 & 49.09 \\ 
& &(\textit{\scriptsize {+2.29}})& (\textit{\scriptsize {+4.80}})&(\textit{\scriptsize {+4.36}})&(\textit{\scriptsize {+1.08}})&(\textit{\scriptsize {+17.05}}) \\ 
DenseDCT\_Swin-T$_{attn}$& 28M   &84.51 & 67.19  & 95.25& 57.59 & 48.21 \\ 
& &(\textit{\scriptsize {+0.31}})& (\textit{\scriptsize {+0.09}})&(\textit{\scriptsize {-0.40}})&(\textit{\scriptsize {+0.23}})&(\textit{\scriptsize {-0.88}}) \\ 
\hline
Focal-Tiny \cite{yang2021focal} & 29M &   88.31 & 71.58&95.97&63.06 & 32.54\\
DenseDCT\_Focal-T & 29M  &    89.59 & 73.94 & 96.13& 64.11  & 49.56\\
&& (\textit{\scriptsize{+1.28}})& (\textit{\scriptsize{+2.36}}) &(\textit{\scriptsize{+0.16}}) &(\textit{\scriptsize {+0.51}}) &(\textit{\scriptsize {+17.02}})\\ 
DenseDCT\_Focal-T$_{attn}$ & 29M  & 89.62 & 73.49& 95.23 & 64.43  & 49.40\\
&& (\textit{\scriptsize{+0.03}})& (\textit{\scriptsize{-0.45}}) &(\textit{\scriptsize{-0.90}}) &(\textit{\scriptsize {+0.32}}) &(\textit{\scriptsize {-0.16}})\\ \bottomrule
\end{tabular}
\label{table:all}
\end{table}
\setlength{\tabcolsep}{1.4pt}

{Among these 5 datasets, CIFAR-10/100 and SVHN datasets have the smallest image size, 32$\times$32, compared with Tiny ImageNet and Flowers-102, holding less raw information, while the image size for Flowers-102 is greater than 500.} We first test our dense DCT token with Swin Transformer \cite{liu2021swin} and Focal Transformer \cite{yang2021focal}. From Table \ref{table:all}, we can see an obvious performance gap between ResNet50 \cite{he2016deep} and vision Transformers, including Deit-S \cite{touvron2021training}, T2T \cite{yuan2021incorporating}, Swin \cite{liu2021swin} and Focal Transformer \cite{yang2021focal} on all datasets. While with the proposed dense DCT input, the gap is significantly reduced and the performance is close to ResNet. For Swin Transformer, our dense DCT input achieves 4.80$\%$ improvement on CIFAR-100 compared with RGB input and up to 17.05$\%$ on the Flowers-102 dataset. We have achieved a 17.02$\%$ increase on Flowers for Focal-Tiny transformer.

In addition, we also add an attention module, Coordinate Attention \cite{hou2021coordinate}, in our experiments  as illustrated in \ref{fig:comparison}. From Table \ref{table:all}, we can still see an increase for some backbones, e.g. Swin Transformer with Dense DCT further increases 0.31$\%$ on CIFAR-10 dataset after adding the attention module and the attention module also adds 0.32$\%$ to Focal Transformer with Dense DCT input on Tiny-ImageNet. However, it may harm the performance sometimes, e.g. also for Focal Transformer on SVHN dataset, it decreases the accuracy by 0.90$\%$. 

\subsection{Ablation Study}
In this section, we check how each component of the proposed approach affects the performance, including the channel selection strategy, attention module on the input frequency domain, and resizing. We further study how resizing influences accuracy.

\setlength{\tabcolsep}{4pt}
\subsubsection{All Components}

This experiment is based on the Swin Transformer backbone on the Tiny ImageNet dataset. According to the results by setting the threshold to their heatmaps, we select 18 channels for our dense DCT representation. The attention module we employed here is Coordinate Attention \cite{hou2021coordinate}.

\begin{table*}[t]
\centering
\caption{Ablation study on Tiny ImageNet with Swin Transformer backbone. For Tiny ImageNet, the original image size is 64. `Attn' here means adding Coordinate Attention on the DCT input directly}
\begin{tabular}{c|c|c|c|c|c|c|c}
\toprule\noalign{\smallskip}
Input & Size& Patch Size & Input size & Channel \# & Resize & Attn  & Top-1 $\uparrow$ ($\%$) \\ 
\noalign{\smallskip}
\hline
RGB   & 224 &4   & 56         & ---        & \checkmark             & ---         & 56.28    \\ \hline
RGB   & 448 &8   & 56         & ---        & \checkmark             & ---         & {55.59}   \\ \hline
DCT   & 64 &8    & 8          & 192 (all)             & ---             & ---         & 38.50   \\ \hline
DCT   & 448 &8    & 56          & 192 (all)   & \checkmark  & ---    & 53.11(\textit{\scriptsize {+14.61}})   \\ \hline
DCT   & 448  &8   & 56   & 24 \cite{xu2020learning}  & \checkmark & --- & 57.02(\textit{\scriptsize {+3.91}})   \\ \hline \hline
DCT   & 448 &8    & 56   & 18           & \checkmark       & ---  & 57.36 (\textit{\scriptsize {+0.34}})  \\ \hline
DCT   & 448 &8   & 56   & 18     & \checkmark     & \checkmark  & 57.59 (\textit{\scriptsize {+0.23}})   \\ 
\bottomrule
\end{tabular}
\label{table:ablation}
\end{table*}

The results are shown in Table \ref{table:ablation}, where `size' in the second column represents the resizing in Figure \ref{fig:comparison} at the beginning. `Input size' and `Channel \#' denote the height (equal to the width) and the number of channels of the input embedding before the Transformer blocks in Figure \ref{fig:comparison} separately. From the results, we can see that the DCT input with all 192 channels and no resizing on the input yields a bad result, 38.50$\%$. However, after resizing, it becomes 53.11$\%$, increased by 14.61$\%$. The channel selection strategy further improves the performance due to the increase in information density. The proposed heatmap-based channel selection achieves 57.36$\%$, 0.34$\%$ greater than the shape-based channel selection strategy in \cite{xu2020learning} with 6 fewer input channels. {To exclude the influence of larger image size between RGB input and DCT input, we also resize the RGB version to 448 and use $8\time8$ patches in the input embedding layer to get the same input size as DCT input, i.e. $56\times 56$, as shown in row 3. However, resizing to 448 decreases the performance from 56.28$\%$ to 55.59$\%$, inferior to the resizing in frequency domain 57.36$\%$. Based on these results, we fix the image size to 224 with patch size 4 when training RGB input versions in Table \ref{table:all}. Finally, the attention module also brings a 0.23\% increase, indicating the potential to add the attention module to the input images.}

Dense DCT holds some interesting properties.
1) No convolution layer or linear layer on the input layer. The combination of DenseDCT input with Transformers or pure MLP networks makes the whole network more explainable, i.e., the whole framework is literately a function of DCT features, instead of any unknown features from convolution layers or the combination of low semantic pixels.
2) Higher semantic information makes it possible to perform attention at the early stage, literally the raw data since DCT transformation is invertible.
3) Taking larger input images with the same backbone as Swin Transformer. The whole network can benefit a lot from larger input images, either from the original larger image to take more information or from resizing as shown above.
However, there is still some space for improvement. For instance, both our heatmap-based method and dynamic/shape-based method in \cite{xu2020learning} take static input channel, while as shown in Figure \ref{fig:GradCAM}, functioning channels vary from the final selected channels when it comes to every single image, e.g. channel 1 for Cb component in Figure \ref{fig:GradCAM} is selected considering the overall performance for the whole dataset according to \ref{fig:sns_ResNet50}, while activated channel 24 is removed. This compromise indicates included redundancy and encourages us to design a customized dynamic channel selection strategy to further improve useful input information density. 

\subsubsection{Effect of Selected Channels}

In this experiment, we show the results by setting different thresholds to the input channel-wise heatmap obtained by training 8$\times$8 DCT full frequency channels, i.e., 192, and making all channels except the candidate one as in Figure \ref{fig:GradCAM}. As shown in Table \ref{table:threshold}, even with channels 18, 21 and 22, which are less than 24 from the shape-based static channel selection strategy in \cite{xu2020learning}, we also achieve higher accuracy, 57.36, 57.11 and 57.03, compared to 57.02 as shown in Table \ref{table:ablation}. Based on the experiment, we choose 18 channels in our implementation of 8$\times$8 DCT transformation, which happen to be the low-frequency channels according to the heatmap. Apart from these, the accuracy for 17 channels is 56.34 while it's 57.36 for 18 channels, increasing 1.02. This vital channel is the DC component for Y component. 

To check what are those deleted frequency channels and how they affect the RGB images, we visualize some images from the Flowers-102 and Tiny-ImageNet as shown in Appendix. From the visualization, we know that most information is kept even after deleting many frequency channels, from 192 channels to 24 or 18, and only some high-frequency details are lost. This can also be found in columns 3 and 5, where removed information forms contours. Besides this, we also checked the effect of blocks where we calculate the GradCAM, classifier, epochs, and datasets as shown in the Appendix. We found that heatmaps from Flowers-102 tend to focus on different higher frequencies compared with heatmaps from Tiny-ImageNet and CIFAR-10.

\begin{table}[t]
\centering
\caption{Effect of setting different threshold in input channel-wise heatmap (8$\times$8)}
\begin{tabular}{c|ccccccccc}
\noalign{\smallskip}
\toprule
\bf{Threshold}   & 0.08& 0.07 & 0.06 & 0.03 & 0.022& 0.02& 0.01 & 0.000& all \\
Channel \#  & 17 & 18 & 21 & 22    & 24   & 25 & 28 & 41 & 192  \\ \hline
Top-1 ($\%$) & 56.34 & \bf{57.36} & 57.11& 57.03 & 56.86 & 57.02 & 56.89 & 56.94 & 53.11    \\
\bottomrule
\end{tabular}
\label{table:threshold}
\end{table}
\setlength{\tabcolsep}{1.4pt}

\subsubsection{Effect of Resizing}

In this section, we analyze the effectiveness of resizing and show the results in Table \ref{table:resize}, where the window size is set to 8 to adapt to all resizes in the table. Since the input tensor before the Transformer blocks of Swin Transformer is associated with the window size in the attention module inside the Transformer blocks, i.e., should be divisible by the window size, and resizing can affect the shape of the input tensor, we fix the window size to 8 and set resize ratio to 1, 2, 4 and 8 in this experiment to control variables and 18 selected channels from previous results. In Table \ref{table:resize}, the first row denotes the resizing ratio compared with the original image 64 on Tiny ImageNet (ratio 1 indicates no resizing).

\begin{table}
\centering
\caption{Analysis of resizing for DCT input and Swin Transformer.}
\setlength{\tabcolsep}{5pt} 
\begin{tabular}{c|cccc}
\toprule
Resize Ratio & 1 & 2 & 4 & 8 \\ \hline
Image Size & 64 & 128 & 256 & 448 \\ \hline
Accuracy & 40.49 &  49.30 & 54.89 & 56.44   \\ \bottomrule
\end{tabular}
\label{table:resize}
\end{table}

\begin{figure}[t]
\centering
\includegraphics[width=0.4\textwidth]{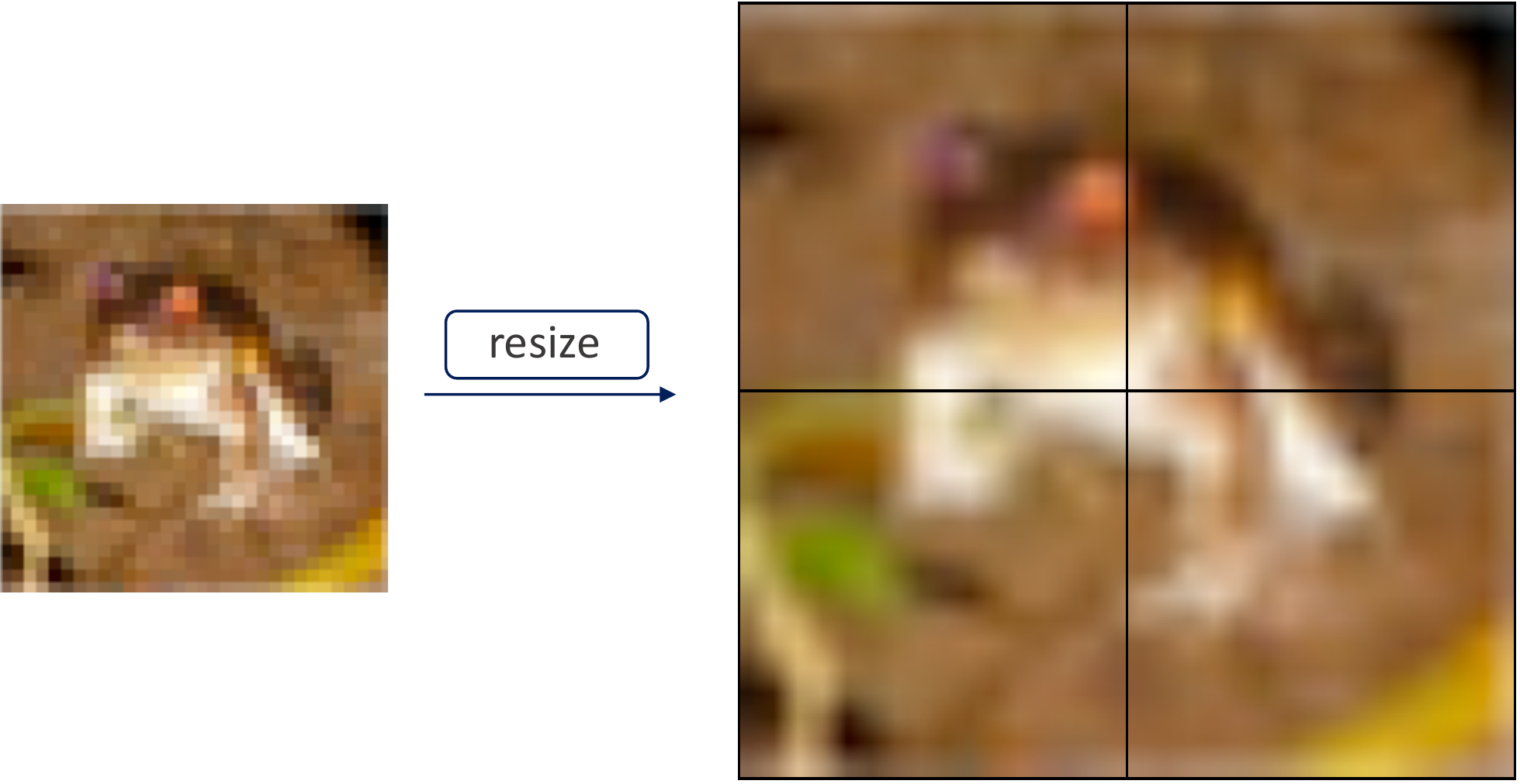}
\caption{The effect of resizing.}
\label{fig:resize}
\end{figure}

From \ref{table:ablation} we can see that resizing brings 14.61$\%$ improvement, from 38.50$\%$ to 53.11$\%$. The same effect is also observed in Table \ref{table:resize}. The accuracy is increased from 40.49$\%$ using original images with size 64 to 56.44$\%$ after resizing to 448. Both results indicate that resizing is a significant preprocessing method for training in the DCT frequency domain. The effect of resizing is illustrated in \ref{fig:resize}. Suppose that the size of the original image is 8$\times$8 before resizing, and an 8$\times$8 DCT is performed on the whole image. While if we resize it to 16$\times$16, the same DCT window only covers one-quarter of the original image, which means sparser features embedded in the image are processed by DCT. The resizing preprocess would better separate useful information from redundant information in the original image and promote further channel selection. Moreover, with smaller raw image blocks after resizing, position information is more accurate and approaches actual positions within each block since block DCT misses position information within blocks. 

{\bf More Analysis: }The appendix provides more analysis of the effect of deleted frequency channels on the RGB images and the influences of channel-wise heatmaps w.r.t. classifiers, blocks, and epochs.

\section{Conclusion}

This paper proposes to explicitly increase input information density to fill the gap between languages and images so as to increase the performance of vision Transformers on small-scale datasets. To achieve this, we have designed a simple yet effective pipeline to calculate the channel-wise heatmaps and select essential frequency channels in the DCT frequency domain. Surprisingly, our parameter-free heatmaps are highly consistent with sampling-based heatmaps from previous work and are more efficient and explainable. It is also simple to implement by using the well-developed JPEG compression API. Classification results on small-scale datasets including CIFAR-10/100, Tiny ImageNet, SVHN, and Flowers-102 show the effectiveness of our approach. Moreover, the selection strategy can be directly implemented in other tasks to reduce the communication cost between GPU and CPU while improving overall performance.

{\bf Acknowledgement.}
This work was partly supported in part by NSERC under grant nos. RGPIN-2021-04244 and ALLRP 576612-22, NIH grant 1R03CA253212-01, USDA grant 2019-67021-28996, NSF CAREER award DBI-1943291, and the Fundamental Research Funds for the Central Universities $2021$RC$244$.

{
\bibliography{reference}

\begin{thebibliography}{10}

\bibitem{cen2021deep}
Feng Cen, Xiaoyu Zhao, Wuzhuang Li, and Guanghui Wang.
\newblock Deep feature augmentation for occluded image classification.
\newblock {\em Pattern Recognition}, 111:107737, 2021.

\bibitem{chenimproving}
Xiangyu Chen, Ying Qin, Wenju Xu, Andr{\'e}s~M Bur, Cuncong Zhong, and Guanghui
  Wang.
\newblock Improving vision transformers on small datasets by increasing input
  information density in frequency domain.
\newblock In {\em IEEE/CVF International Conference on Computer Vision
  Workshops (ICCVW)}, 2022.

\bibitem{deng2009imagenet}
Jia Deng, Wei Dong, Richard Socher, Li-Jia Li, Kai Li, and Li~Fei-Fei.
\newblock Imagenet: A large-scale hierarchical image database.
\newblock In {\em 2009 IEEE conference on computer vision and pattern
  recognition}, pages 248--255. Ieee, 2009.

\bibitem{dosovitskiy2020image}
Alexey Dosovitskiy, Lucas Beyer, Alexander Kolesnikov, Dirk Weissenborn,
  Xiaohua Zhai, Thomas Unterthiner, Mostafa Dehghani, Matthias Minderer, Georg
  Heigold, Sylvain Gelly, et~al.
\newblock An image is worth 16x16 words: Transformers for image recognition at
  scale.
\newblock In {\em ICLR}, 2021.

\bibitem{ehrlich2019deep}
Max Ehrlich and Larry~S Davis.
\newblock Deep residual learning in the jpeg transform domain.
\newblock In {\em Proceedings of the IEEE/CVF International Conference on
  Computer Vision}, pages 3484--3493, 2019.

\bibitem{fan2021multiscale}
Haoqi Fan, Bo~Xiong, Karttikeya Mangalam, Yanghao Li, Zhicheng Yan, Jitendra
  Malik, and Christoph Feichtenhofer.
\newblock Multiscale vision transformers.
\newblock In {\em Proceedings of the IEEE/CVF International Conference on
  Computer Vision}, pages 6824--6835, 2021.

\bibitem{gajurel2021fine}
Kamala Gajurel, Cuncong Zhong, and Guanghui Wang.
\newblock A fine-grained visual attention approach for fingerspelling
  recognition in the wild.
\newblock In {\em 2021 IEEE International Conference on Systems, Man, and
  Cybernetics (SMC)}, pages 3266--3271. IEEE, 2021.

\bibitem{gani2022train}
Hanan Gani, Muzammal Naseer, and Mohammad Yaqub.
\newblock How to train vision transformer on small-scale datasets?
\newblock {\em arXiv preprint arXiv:2210.07240}, 2022.

\bibitem{ghosh2016deep}
Arthita Ghosh and Rama Chellappa.
\newblock Deep feature extraction in the dct domain.
\newblock In {\em 2016 23rd International Conference on Pattern Recognition
  (ICPR)}, pages 3536--3541. IEEE, 2016.

\bibitem{graham2021levit}
Benjamin Graham, Alaaeldin El-Nouby, Hugo Touvron, Pierre Stock, Armand Joulin,
  Herv{\'e} J{\'e}gou, and Matthijs Douze.
\newblock Levit: a vision transformer in convnet's clothing for faster
  inference.
\newblock In {\em Proceedings of the IEEE/CVF International Conference on
  Computer Vision}, pages 12259--12269, 2021.

\bibitem{gueguen2018faster}
Lionel Gueguen, Alex Sergeev, Ben Kadlec, Rosanne Liu, and Jason Yosinski.
\newblock Faster neural networks straight from jpeg.
\newblock {\em Advances in Neural Information Processing Systems}, 31, 2018.

\bibitem{guo2021cmt}
Jianyuan Guo, Kai Han, Han Wu, Chang Xu, Yehui Tang, Chunjing Xu, and Yunhe
  Wang.
\newblock Cmt: Convolutional neural networks meet vision transformers.
\newblock {\em arXiv preprint arXiv:2107.06263}, 2021.

\bibitem{he2022masked}
Kaiming He, Xinlei Chen, Saining Xie, Yanghao Li, Piotr Doll{\'a}r, and Ross
  Girshick.
\newblock Masked autoencoders are scalable vision learners.
\newblock In {\em Proceedings of the IEEE/CVF International Conference on
  Computer Vision}, 2022.

\bibitem{he2016deep}
Kaiming He, Xiangyu Zhang, Shaoqing Ren, and Jian Sun.
\newblock Deep residual learning for image recognition.
\newblock In {\em Proceedings of the IEEE conference on computer vision and
  pattern recognition}, pages 770--778, 2016.

\bibitem{hou2021coordinate}
Qibin Hou, Daquan Zhou, and Jiashi Feng.
\newblock Coordinate attention for efficient mobile network design.
\newblock In {\em Proceedings of the IEEE/CVF Conference on Computer Vision and
  Pattern Recognition}, pages 13713--13722, 2021.

\bibitem{krizhevsky2009learning}
Alex Krizhevsky, Geoffrey Hinton, et~al.
\newblock Learning multiple layers of features from tiny images.
\newblock 2009.

\bibitem{le2015tiny}
Ya~Le and Xuan Yang.
\newblock Tiny imagenet visual recognition challenge.
\newblock {\em CS 231N}, 7(7):3, 2015.

\bibitem{lee2021vision}
Seung~Hoon Lee, Seunghyun Lee, and Byung~Cheol Song.
\newblock Vision transformer for small-size datasets.
\newblock {\em arXiv preprint arXiv:2112.13492}, 2021.

\bibitem{li2021localvit}
Yawei Li, Kai Zhang, Jiezhang Cao, Radu Timofte, and Luc Van~Gool.
\newblock Localvit: Bringing locality to vision transformers.
\newblock {\em arXiv preprint arXiv:2104.05707}, 2021.

\bibitem{liu2021efficient}
Yahui Liu, Enver Sangineto, Wei Bi, Nicu Sebe, Bruno Lepri, and Marco Nadai.
\newblock Efficient training of visual transformers with small datasets.
\newblock {\em Advances in Neural Information Processing Systems}, 34, 2021.

\bibitem{liu2021swin}
Ze~Liu, Yutong Lin, Yue Cao, Han Hu, Yixuan Wei, Zheng Zhang, Stephen Lin, and
  Baining Guo.
\newblock Swin transformer: Hierarchical vision transformer using shifted
  windows.
\newblock In {\em Proceedings of the IEEE/CVF International Conference on
  Computer Vision}, pages 10012--10022, 2021.

\bibitem{ma2021miti}
Wenchi Ma, Tianxiao Zhang, and Guanghui Wang.
\newblock Miti-detr: Object detection based on transformers with mitigatory
  self-attention convergence.
\newblock {\em arXiv preprint arXiv:2112.13310}, 2021.

\bibitem{netzer2011reading}
Yuval Netzer, Tao Wang, Adam Coates, Alessandro Bissacco, Bo~Wu, and Andrew~Y
  Ng.
\newblock Reading digits in natural images with unsupervised feature learning.
\newblock 2011.

\bibitem{Nilsback08}
Maria-Elena Nilsback and Andrew Zisserman.
\newblock Automated flower classification over a large number of classes.
\newblock In {\em Indian Conference on Computer Vision, Graphics and Image
  Processing}, Dec 2008.

\bibitem{pan2021less}
Zizheng Pan, Bohan Zhuang, Haoyu He, Jing Liu, and Jianfei Cai.
\newblock Less is more: Pay less attention in vision transformers.
\newblock In {\em AAAI Conference on Artificial Intelligence (AAAI), 2022}.

\bibitem{patel2022aggregating}
Krushi Patel, Andres~M Bur, Fengjun Li, and Guanghui Wang.
\newblock Aggregating global features into local vision transformer.
\newblock {\em arXiv preprint arXiv:2201.12903}, 2022.

\bibitem{patel2022discriminative}
Krushi Patel and Guanghui Wang.
\newblock A discriminative channel diversification network for image
  classification.
\newblock {\em Pattern Recognition Letters}, 153:176--182, 2022.

\bibitem{sajid2021audio}
Usman Sajid, Xiangyu Chen, Hasan Sajid, Taejoon Kim, and Guanghui Wang.
\newblock Audio-visual transformer based crowd counting.
\newblock In {\em Proceedings of the IEEE/CVF International Conference on
  Computer Vision}, pages 2249--2259, 2021.

\bibitem{selvaraju2017grad}
Ramprasaath~R Selvaraju, Michael Cogswell, Abhishek Das, Ramakrishna Vedantam,
  Devi Parikh, and Dhruv Batra.
\newblock Grad-cam: Visual explanations from deep networks via gradient-based
  localization.
\newblock In {\em Proceedings of the IEEE international conference on computer
  vision}, pages 618--626, 2017.

\bibitem{sun2017revisiting}
Chen Sun, Abhinav Shrivastava, Saurabh Singh, and Abhinav Gupta.
\newblock Revisiting unreasonable effectiveness of data in deep learning era.
\newblock In {\em Proceedings of the IEEE international conference on computer
  vision}, pages 843--852, 2017.

\bibitem{touvron2021training}
Hugo Touvron, Matthieu Cord, Matthijs Douze, Francisco Massa, Alexandre
  Sablayrolles, and Herv{\'e} J{\'e}gou.
\newblock Training data-efficient image transformers \& distillation through
  attention.
\newblock In {\em International Conference on Machine Learning}, pages
  10347--10357. PMLR, 2021.

\bibitem{wang2021pyramid}
Wenhai Wang, Enze Xie, Xiang Li, Deng-Ping Fan, Kaitao Song, Ding Liang, Tong
  Lu, Ping Luo, and Ling Shao.
\newblock Pyramid vision transformer: A versatile backbone for dense prediction
  without convolutions.
\newblock In {\em Proceedings of the IEEE/CVF International Conference on
  Computer Vision}, pages 568--578, 2021.

\bibitem{wu2021cvt}
Haiping Wu, Bin Xiao, Noel Codella, Mengchen Liu, Xiyang Dai, Lu~Yuan, and Lei
  Zhang.
\newblock Cvt: Introducing convolutions to vision transformers.
\newblock In {\em Proceedings of the IEEE/CVF International Conference on
  Computer Vision}, pages 22--31, 2021.

\bibitem{xiao2021early}
Tete Xiao, Piotr Dollar, Mannat Singh, Eric Mintun, Trevor Darrell, and Ross
  Girshick.
\newblock Early convolutions help transformers see better.
\newblock {\em Advances in Neural Information Processing Systems}, 34, 2021.

\bibitem{xu2020learning}
Kai Xu, Minghai Qin, Fei Sun, Yuhao Wang, Yen-Kuang Chen, and Fengbo Ren.
\newblock Learning in the frequency domain.
\newblock In {\em Proceedings of the IEEE/CVF Conference on Computer Vision and
  Pattern Recognition}, pages 1740--1749, 2020.

\bibitem{xu2021co}
Weijian Xu, Yifan Xu, Tyler Chang, and Zhuowen Tu.
\newblock Co-scale conv-attentional image transformers.
\newblock In {\em Proceedings of the IEEE/CVF International Conference on
  Computer Vision}, pages 9981--9990, 2021.

\bibitem{yang2021focal}
Jianwei Yang, Chunyuan Li, Pengchuan Zhang, Xiyang Dai, Bin Xiao, Lu~Yuan, and
  Jianfeng Gao.
\newblock Focal self-attention for local-global interactions in vision
  transformers.
\newblock {\em arXiv preprint arXiv:2107.00641}, 2021.

\bibitem{yang2022unsupervised}
Yiju Yang, Tianxiao Zhang, Guanyu Li, Taejoon Kim, and Guanghui Wang.
\newblock An unsupervised domain adaptation model based on dual-module
  adversarial training.
\newblock {\em Neurocomputing}, 475:102--111, 2022.

\bibitem{yuan2021incorporating}
Kun Yuan, Shaopeng Guo, Ziwei Liu, Aojun Zhou, Fengwei Yu, and Wei Wu.
\newblock Incorporating convolution designs into visual transformers.
\newblock In {\em Proceedings of the IEEE/CVF International Conference on
  Computer Vision}, pages 579--588, 2021.

\bibitem{yuan2021tokens}
Li~Yuan, Yunpeng Chen, Tao Wang, Weihao Yu, Yujun Shi, Zi-Hang Jiang,
  Francis~EH Tay, Jiashi Feng, and Shuicheng Yan.
\newblock Tokens-to-token vit: Training vision transformers from scratch on
  imagenet.
\newblock In {\em Proceedings of the IEEE/CVF International Conference on
  Computer Vision}, pages 558--567, 2021.

\bibitem{yun2019cutmix}
Sangdoo Yun, Dongyoon Han, Seong~Joon Oh, Sanghyuk Chun, Junsuk Choe, and
  Youngjoon Yoo.
\newblock Cutmix: Regularization strategy to train strong classifiers with
  localizable features.
\newblock In {\em Proceedings of the IEEE/CVF international conference on
  computer vision}, pages 6023--6032, 2019.

\bibitem{zhang2017mixup}
Hongyi Zhang, Moustapha Cisse, Yann~N Dauphin, and David Lopez-Paz.
\newblock mixup: Beyond empirical risk minimization.
\newblock {\em arXiv preprint arXiv:1710.09412}, 2017.

\bibitem{zhang2022nested}
Zizhao Zhang, Han Zhang, Long Zhao, Ting Chen, Sercan Arik, and Tomas Pfister.
\newblock Nested hierarchical transformer: Towards accurate, data-efficient and
  interpretable visual understanding.
\newblock In {\em AAAI Conference on Artificial Intelligence (AAAI), 2022}.

\end{thebibliography}
\bibliographystyle{plain}
}

\newpage
\appendix
\section{Appendix}

\subsection{Visualization}
\begin{figure}[ht]
\centering
\includegraphics[width=\linewidth, height=2.5cm]{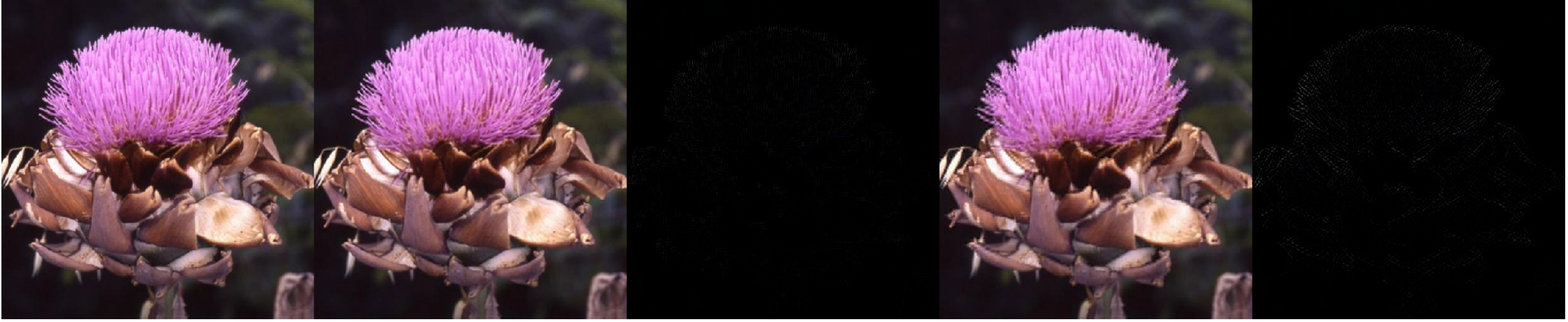}
\includegraphics[width=\linewidth, height=2.5cm]{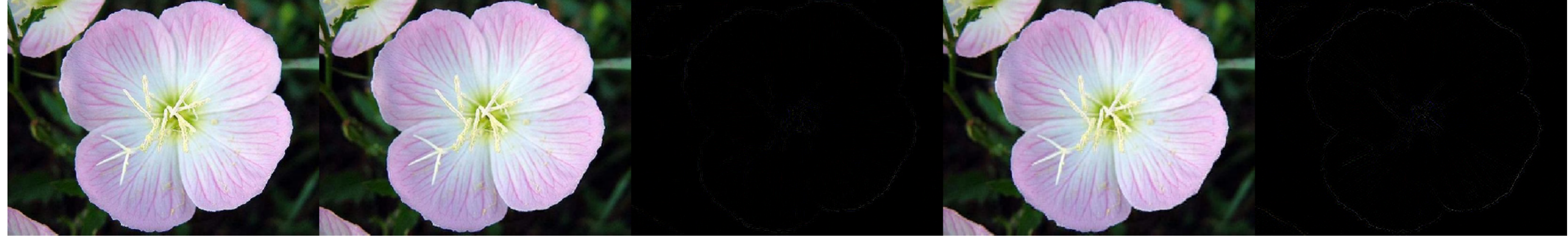}
\includegraphics[width=\linewidth, height=2.5cm]{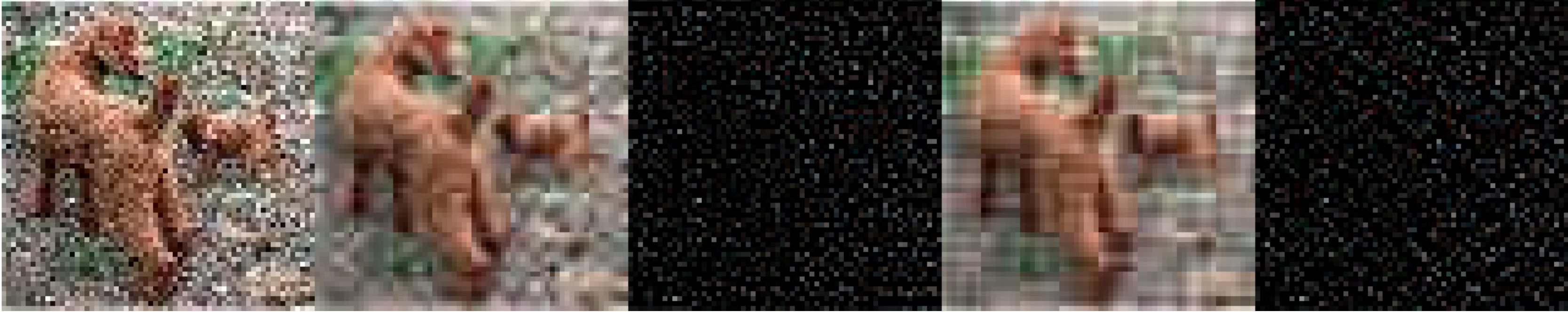}
\includegraphics[width=\linewidth, height=2.5cm]{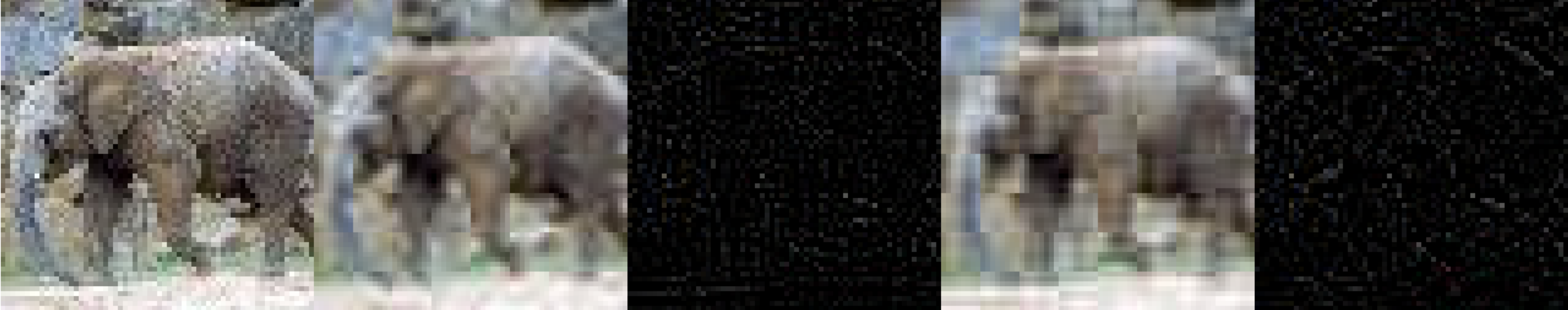}
\includegraphics[width=\linewidth, height=2.5cm]{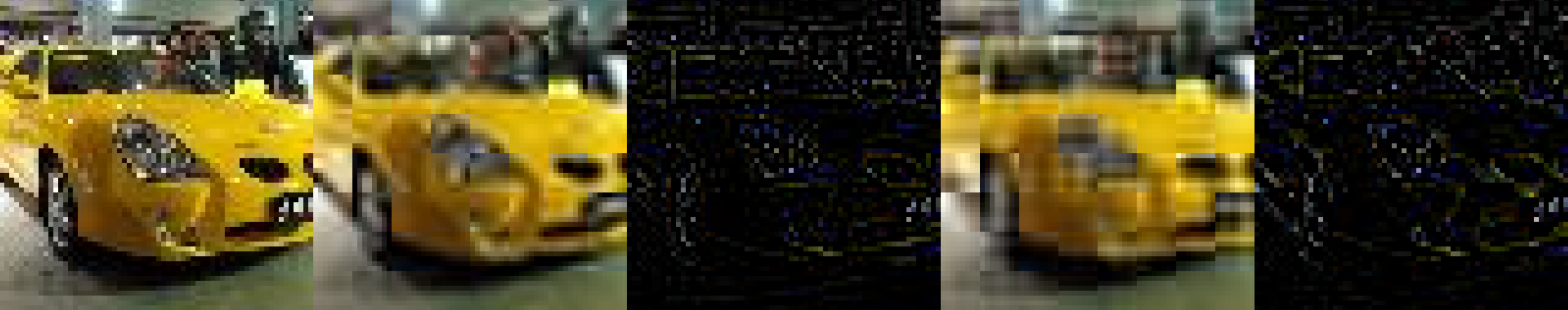}
\includegraphics[width=\linewidth, height=2.5cm]{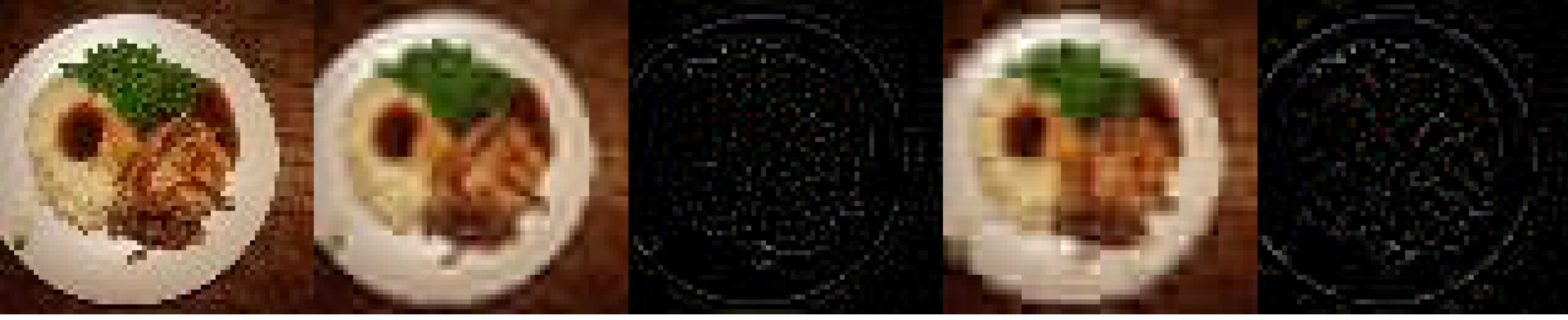}
\caption{Visualization of the effect of removing frequency channels on RGB domain. From left to right column: (1) original image; (2) IDCT$_{24}$: reconstructed image after removing frequency channels as in [27] and keeping only 24 channels; (3) image-IDCT$_{24}$: removed information in RGB domain; (4) IDCT$_{18}$: reconstructed image with our channel selection strategy; (5) image-IDCT$_{18}$: our removed information in RGB domain. Images from the first 2 rows are from Flowers-102 dataset and the rest rows are from Tiny-ImageNet dataset.}
\label{fig:idct}
\end{figure}

{To check what are those deleted frequency channels and how they affect the RGB images, this section visualizes some images from the Flowers-102 (first 2 rows) and Tiny-ImageNet (since the third row) as shown in Figure \ref{fig:idct}. The first column represents the original images, IDCT$_{24}$ and  IDCT$_{18}$ denote the inverse DCT results of the final kept frequency representations in Figure 2 in Section 3.1. From columns 2 and 4 we know, that most information is kept even after deleting many frequency channels, from 192 channels to 24 or 18, and only some high-frequency details are lost. This can also be found in columns 3 and 5, where removed information forms contours.}

\subsection{A Closer Look at Channel-Wise Heatmaps}
This section explores what influences the channel-wise heatmaps and how they change them. All experiments are based on 100-epoch training and heatmaps in this section are calculated on 256 images (1 batch) for demonstration.

We first checked the effect of backbones. Following the same setting as in Figure 3 in Section 3.2, ResNet50 and Deit-S are trained. ResNet50 returns a concentrated heatmap as in Figure 5 in Section 3.2 while Deit-S heatmaps are scattered. Based on this, we used ResNet50 for all channel selections. In this sense, knowledge learned from ResNet50 is transferred to vision Transformers in Table 2.

\subsubsection{Cosine Classifier vs Linear Classifier}
\begin{figure}[ht]
\centering
\includegraphics[width=1.0\linewidth]{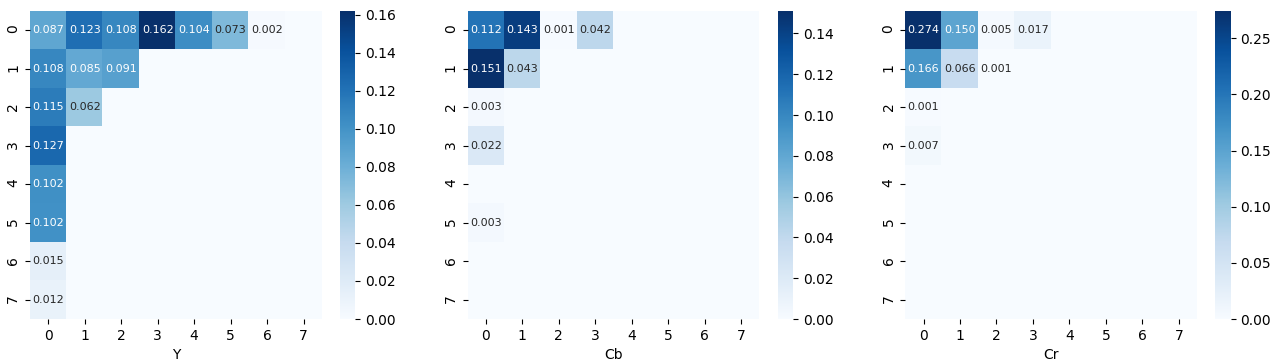}
\centering
\includegraphics[width=1.0\linewidth]{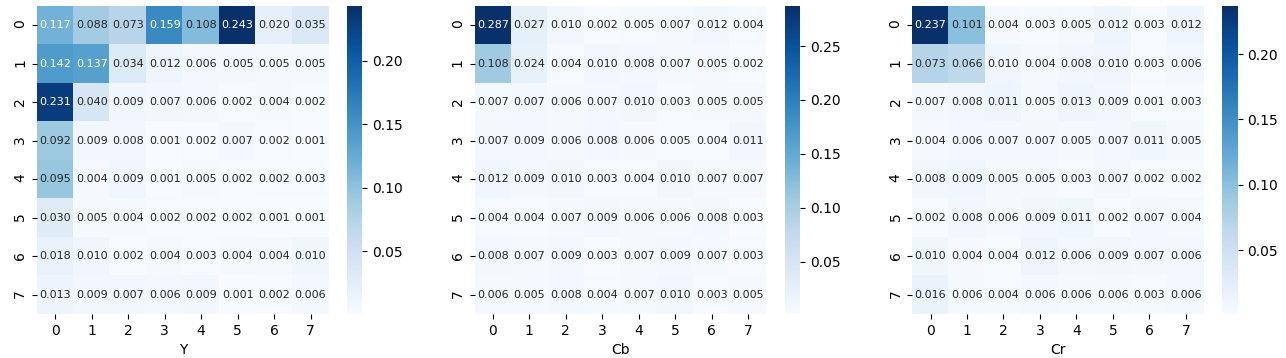}
\caption{Channel-wise heatmaps with different classifier head. Upper: cosine classifier. Lower: linear classifier.}
\label{fig:head}
\end{figure}

In the general classification framework, images are passed into backbones like ResNet50, followed by a linear classifier and softmax function to obtain classification results. While as shown in Figure 3 in Section 3.2, to get the pre-trained model for channel-wise heatmaps, frequency representations after block DCT are fed into backbones like ResNet50, and the following is a cosine classifier instead of a linear classifier as often used in few-shot learning algorithms [1]. To examine the difference, the same channel-wise heatmap generating process is conducted based on ResNet50. Resulted heatmaps are as Figure \ref{fig:head}. As we can see, both heatmaps show similar preferences on frequency channels and most of them lie in the top left corners. As the heatmap from the cosine head is more similar to the one in [27] by sampling, we choose the cosine head to obtain heatmaps everywhere else in this paper.

\subsubsection{Heatmaps on Different Blocks}
\begin{figure}[ht]
\centering
      \includegraphics[width=\linewidth, height=2.5cm]{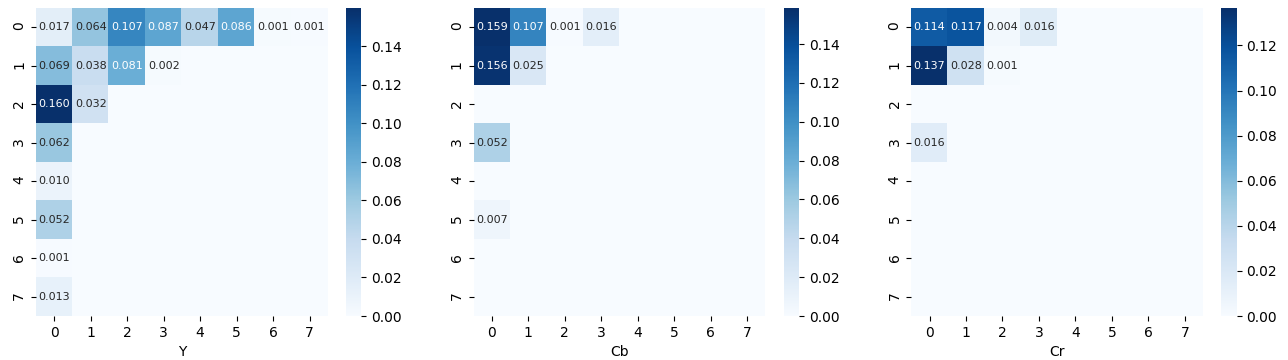}
      \includegraphics[width=\linewidth, height=2.5cm]{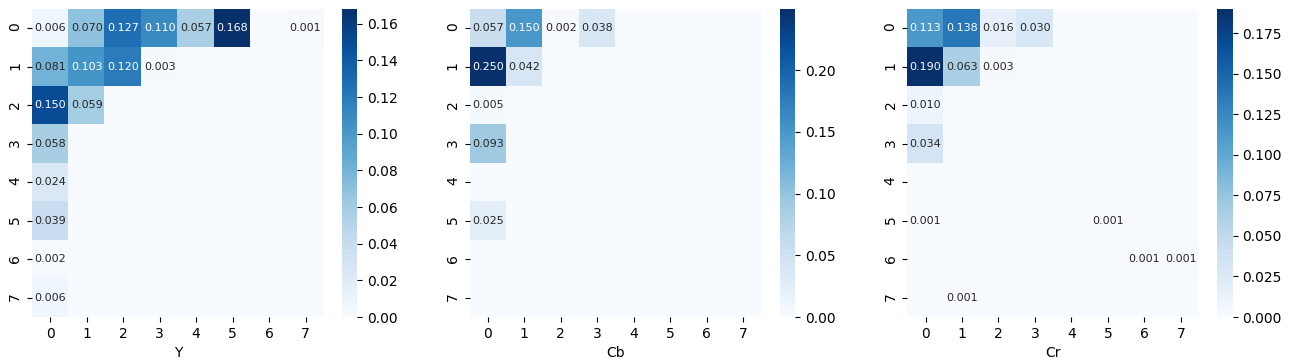}
      \includegraphics[width=\linewidth, height=2.5cm]{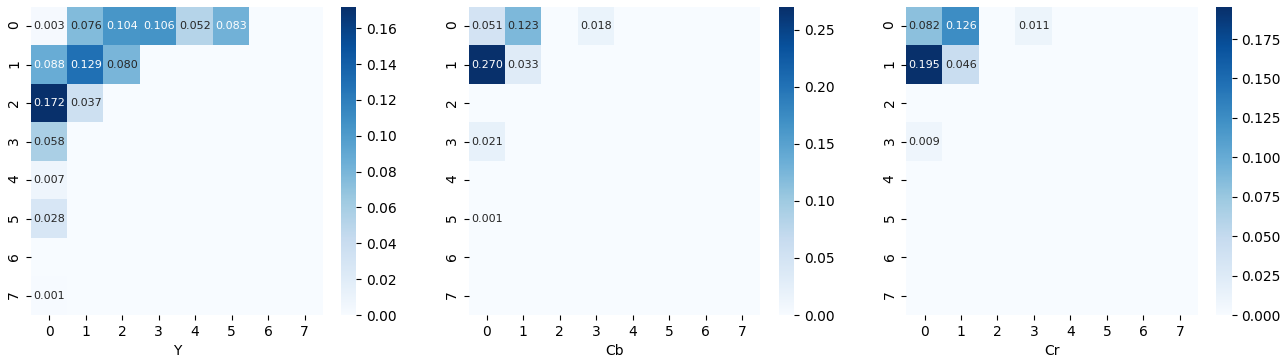}
      \includegraphics[width=\linewidth, height=2.5cm]{./images/heatmaps/figlinear/resnet50_tiny_cosine_best.png}
\caption{Heatmaps corresponding to different blocks in ResNet50. From top to bottom are from first four blocks, block 1 to block 4.}
\label{fig:heatmap_block}
\end{figure}

This section checks the heatmaps corresponding to different blocks in ResNet50 on the Tiny-ImageNet dataset. The best model after 100 epochs is used. From Figure \ref{fig:heatmap_block}, the first 4 layers focus only on the low-frequency channels for the Y, Cb, and Cr components. After that, higher frequency channels are emphasized lightly like block 6, 7, and 12, and heavily like block 5 and 8. We simply use the heatmap from block 4 for qualitative tables in previous sections.

\subsubsection{Heatmaps Across Epochs}
\begin{figure}[ht]
\centering
      \includegraphics[width=\linewidth, height=2.5cm]{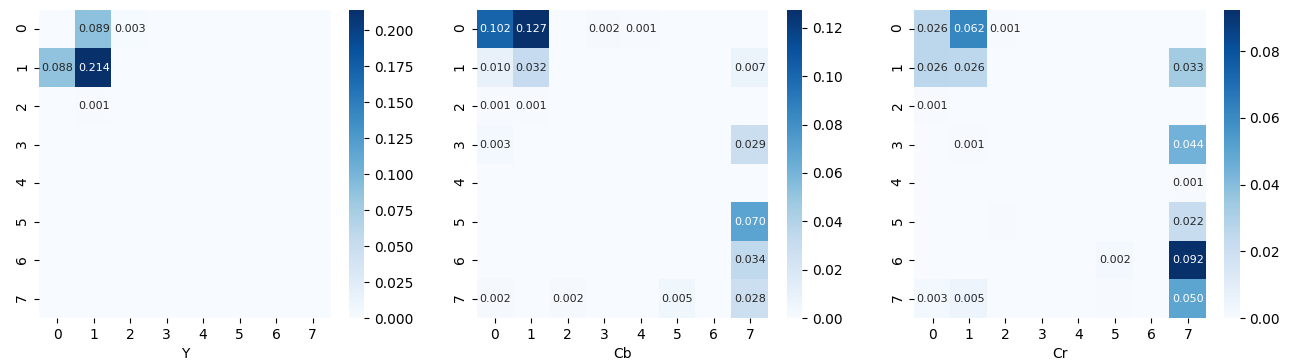}
      \includegraphics[width=\linewidth, height=2.5cm]{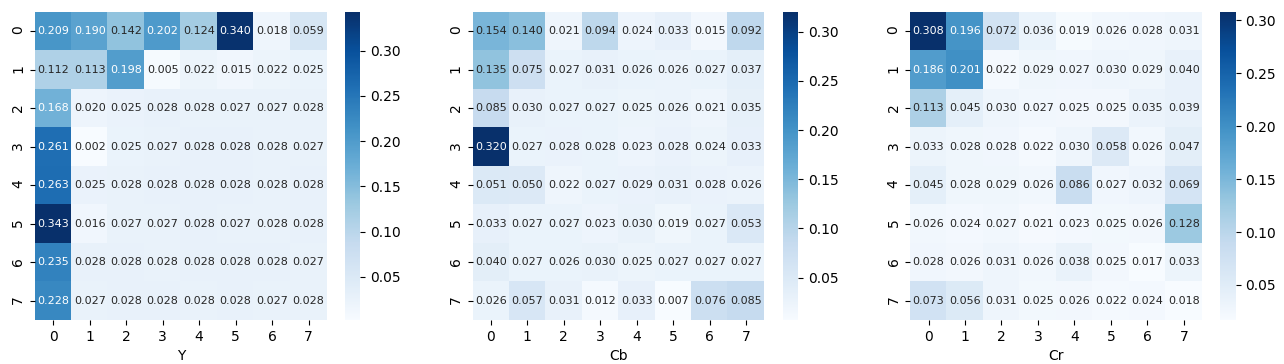}
      \includegraphics[width=\linewidth, height=2.5cm]{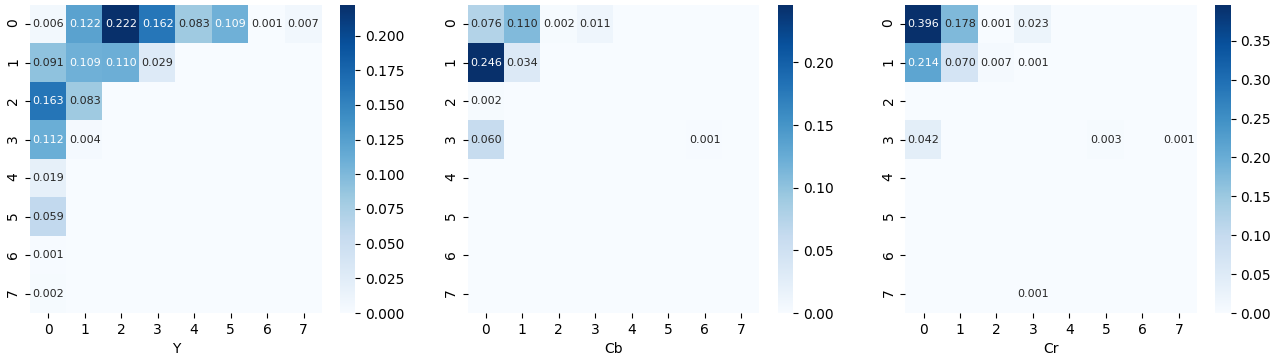}
      \includegraphics[width=\linewidth, height=2.5cm]{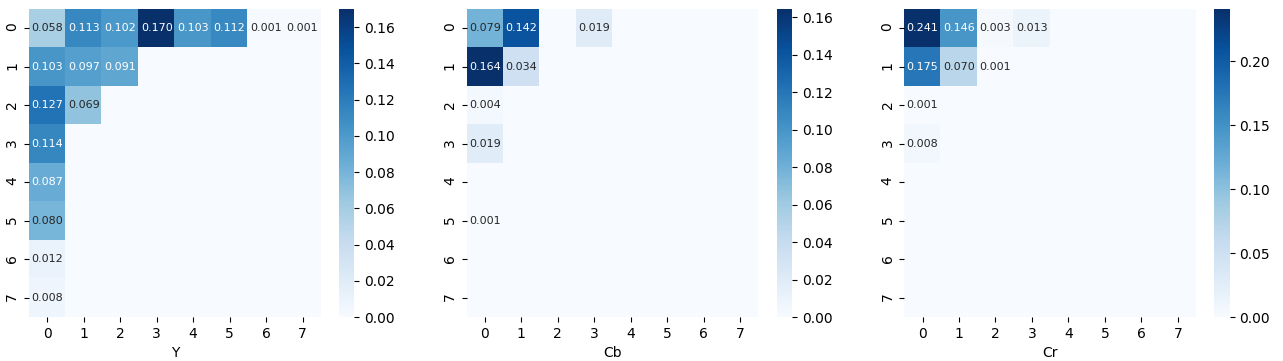}
      \includegraphics[width=\linewidth, height=2.5cm]{./images/heatmaps/figlinear/resnet50_tiny_cosine_best.png}
\caption{Heatmaps corresponding to different epochs in ResNet50. From top to bottom, epoch 0, 30, 60, 90 and best (epoch 98).}
\label{fig:heatmap_epoch}
\end{figure}

This section examines the growth of heatmaps across epochs, including epochs 0, 30, 60, 90 and 98 (when we get the best model within 100 epochs). From Figure \ref{fig:heatmap_epoch}, the network focuses on sparse frequency channels at the beginning, and then it explores more high-frequency channels at epoch 30. Starting from epoch 60, frequency channels are determined focusing on finetuning. It becomes closer and closer to the heatmap from the best model, which is used to calculate the heatmaps throughout this paper. 

\subsubsection{The Effect of Datasets}
\begin{figure}[ht]
\centering
      \includegraphics[width=\linewidth, height=2.5cm]{./images/heatmaps/blocks/b3_resnet50_tiny_cosine_best.png}
      \includegraphics[width=\linewidth, height=2.5cm]{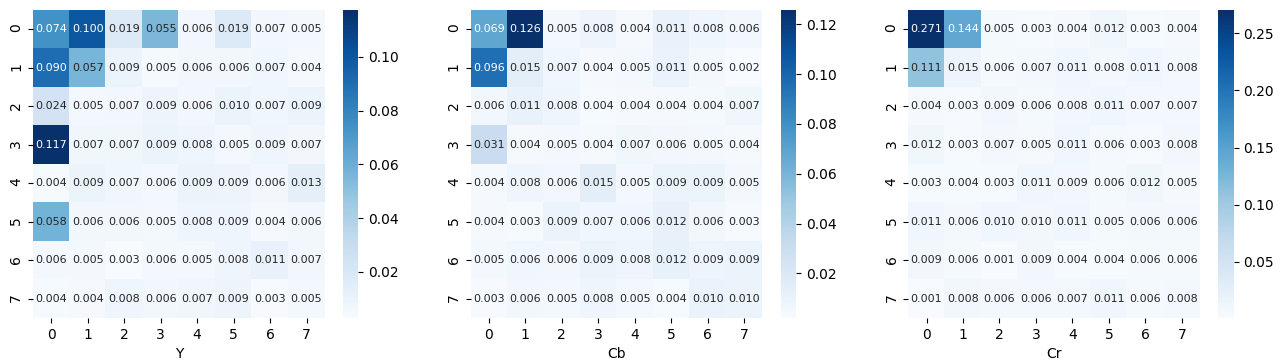}
      \includegraphics[width=\linewidth, height=2.5cm]{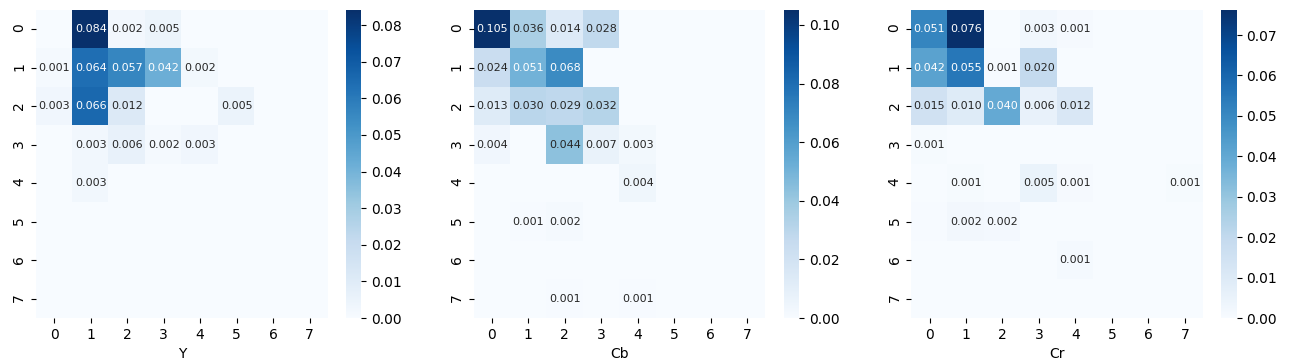}
\caption{Heatmaps correspond to block 3 on different datasets with backbone ResNet50. From top to bottom, Tiny-ImageNet, CIFAR-10 and Flowers-102 respectively.}
\label{fig:heatmap_dataset}
\end{figure}

This section shows the heatmaps of block 3 on different datasets, Tiny-ImageNet, CIFAR-10, and Flowers-102. From Figure \ref{fig:heatmap_dataset}, Tiny-ImageNet and CIFAR-10 focus more on similar frequency channels, while Flowers-102 on a different one. This is because Tiny-ImageNet and CIFAR-10 contain similar categories since CIFAR-10 is a subset of Tiny-ImageNet while images in Flowers-102 are more fine-grained, including only 102 classes of flowers. Moreover, from Table 2, the accuracy on the Flowers-102 dataset is relatively small compared with other datasets, indicating the selected channels are far from reliable convergence. 

\end{document}